\documentclass[10pt,twocolumn,letterpaper]{article}

\usepackage[pagenumbers]{cvpr} 

\usepackage{graphicx}
\usepackage{amsmath}
\usepackage{amssymb}
\usepackage{booktabs}

\usepackage{multirow}
\usepackage{makecell}
\usepackage{graphicx}
\usepackage{enumitem}
\usepackage{comment}
\usepackage{listings}
\setlist{noitemsep,leftmargin=*}
\DeclareMathOperator{\atantwo}{atan2}
\usepackage{wrapfig}
\usepackage{subcaption} 

\newcommand\blfootnote[1]{%
\begingroup
\renewcommand\thefootnote{}\footnote{#1}%
\addtocounter{footnote}{-1}%
\endgroup
}

\usepackage[pagebackref,breaklinks,colorlinks]{hyperref}

\usepackage[capitalize]{cleveref}
\crefname{section}{Sec.}{Secs.}
\Crefname{section}{Section}{Sections}
\Crefname{table}{Table}{Tables}
\crefname{table}{Tab.}{Tabs.}

\def\cvprPaperID{922} 
\def\confName{CVPR}
\def\confYear{2023}

\begin{document}

\title{PVT-SSD: Single-Stage 3D Object Detector with Point-Voxel Transformer}

\author{
Honghui Yang\textsuperscript{\rm 1,3$*$}~~~~
Wenxiao Wang\textsuperscript{\rm 2}~~~~
Minghao Chen\textsuperscript{\rm 1}~~~~
Binbin Lin\textsuperscript{\rm 2$\dag$}~~~~
Tong He\textsuperscript{\rm 3$\dag$}~~~~ \\
Hua Chen\textsuperscript{\rm 4}~~~~
Xiaofei He\textsuperscript{\rm 1}~~~~
Wanli Ouyang\textsuperscript{\rm 3} \\
\textsuperscript{\rm 1}State Key Lab of CAD\&CG, Zhejiang University \\
\textsuperscript{\rm 2}School of Software Technology, Zhejiang University \\
\textsuperscript{\rm 3}Shanghai AI Laboratory \\
\textsuperscript{\rm 4}COMAC Beijing Aircraft Technology Research Institute\\
}
\maketitle

\begin{abstract}
Recent Transformer-based 3D object detectors learn point cloud features either from point- or voxel-based representations.
However, the former requires time-consuming sampling while the latter introduces quantization errors. 
In this paper, we present a novel Point-Voxel Transformer for single-stage 3D detection (PVT-SSD) that takes advantage of these two representations.
Specifically, we first use voxel-based sparse convolutions for efficient feature encoding.
Then, we propose a Point-Voxel Transformer (PVT) module that obtains long-range contexts in a cheap manner from voxels while attaining accurate positions from points.
The key to associating the two different representations is our introduced input-dependent Query Initialization module, which could efficiently generate reference points and content queries.
Then, PVT adaptively fuses long-range contextual and local geometric information around reference points into content queries.
Further, to quickly find the neighboring points of reference points, we design the Virtual Range Image module, which generalizes the native range image to multi-sensor and multi-frame.
The experiments on several autonomous driving benchmarks verify the effectiveness and efficiency of the proposed method.
Code will be available at \url{https://github.com/Nightmare-n/PVT-SSD}.
\blfootnote{$^*$This work was done when Honghui was an intern at Shanghai Artificial Intelligence Laboratory.}
\blfootnote{$^\dag$Corresponding author}
\end{abstract}
\section{Introduction}
3D object detection from point clouds has become increasingly popular thanks to its wide applications, e.g., autonomous driving and virtual reality.
To process unordered point clouds, Transformer~\cite{ashish2017transformer} has recently attracted great interest as the self-attention is invariant to the permutation of inputs.
However, due to the quadratic complexity of self-attention, it involves extensive computation and memory budgets when processing large point clouds.
To overcome this problem, some point-based methods~\cite{misra20213detr,liu2021groupfree,pan2021pointformer} perform attention on downsampled point sets, while some voxel-based methods~\cite{mao2021votr,fan2022sst,yang2022eqpvrcnn} employ attention on local non-empty voxels.
Nevertheless, the former requires farthest point sampling (FPS)~\cite{qi2017pointnet++} to sample point clouds, which is time-consuming on large-scale outdoor scenes~\cite{hu2020randlanet}, while the latter inevitably introduces quantization errors during voxelization, which loses accurate position information.

In this paper, we propose PVT-SSD that absorbs the advantages of the above two representations, i.e., voxels and points, while overcoming their drawbacks.
To this end, instead of sampling points directly, we convert points to a small number of voxels through sparse convolutions and sample non-empty voxels to reduce the runtime of FPS.
Then, inside the PVT-SSD, voxel features are adaptively fused with point features to make up for the quantization error.
In this way, both long-range contexts from voxels and accurate positions from points are preserved. 
Specifically, PVT-SSD consists of the following components:


Firstly, we propose an input-dependent \textbf{Query Initialization} module inspired by previous indoor Transformer-based detectors~\cite{misra20213detr,liu2021groupfree}, which provides queries with better initial positions and instance-related features.
Unlike~\cite{misra20213detr,liu2021groupfree}, our queries originate from non-empty voxels instead of points to reduce the sampling time.
Concretely, with the 3D voxels generated by sparse convolutions, we first \textit{collapse} 3D voxels into 2D voxels by merging voxels along the height dimension to further reduce the number of voxels.
The \textit{sample} operation is then applied to select a representative set of voxels.
We finally \textit{lift} sampled 2D voxels to generate 3D reference points.
Subsequently, the corresponding content queries are obtained in an efficient way by projecting reference points onto a BEV feature map and indexing features at the projected locations.

Secondly, we introduce a \textbf{Point-Voxel Transformer} module that captures long-range contextual features from voxel tokens and extracts fine-grained point features from point tokens.
To be specific, the voxel tokens are obtained from non-empty voxels around reference points to cover a large attention range.
In contrast, the point tokens are generated from neighboring points near reference points to retain fine-grained information.
These two different tokens are adaptively fused by the cross-attention layer based on the similarity with content queries to complement each other.

Furthermore, we design a \textbf{Virtual Range Image} module to accelerate the neighbor querying process in the point-voxel Transformer.
With the constructed range image, reference points can quickly find their neighbors based on range image coordinates.
Unlike the native range image captured by LiDAR sensors, we can handle situations where multiple points overlap on the same pixel in the range image. Therefore, it can be used for complex scenarios, such as multiple sensors and multi-frame fusion.

Extensive experiments have been conducted on several detection benchmarks to verify the efficacy and efficiency of our approach.
PVT-SSD achieves competitive results on KITTI~\cite{geiger2012kitti}, Waymo Open Dataset~\cite{sun2020wod}, and nuScenes~\cite{caesar2020nuscenes}.

\section{Related Work}
\textbf{3D Object Detection from Point Clouds.}
Current 3D detectors can be mainly divided into voxel-, point-, and point-voxel-based methods.
To process irregular 3D point clouds, voxel-based methods~\cite{zheng2020ciassd,zheng2021sessd,he2020sassd,wang2020piilar-od,Yang2018pixor,deng2021voxelrcnn,wu2022sfd} project them onto regular voxels.
VoxelNet~\cite{zhou2018voxelnet} leverages PointNet~\cite{qi2017pointnet} to generate a voxel-wise representation and applies standard 3D and 2D convolutions for object detection. 
PointPillars~\cite{lang2019pointpillar} simplifies the voxels to pillars.
CenterPoint~\cite{yin2021center} estimates the centers of objects using a keypoint detector and removes the need for axis-aligned anchor boxes.
Though voxel-based methods achieve good detection performance with promising efficiency, voxelization inevitably introduces quantization errors.

Point-based methods~\cite{shi2020pointgnn,shi2019pointrcnn,shi2020part,yang2019std} overcome this by directly operating on raw point clouds.
VoteNet~\cite{qi2019votenet} detects 3D objects through Hough voting and clustering.
3DSSD~\cite{yang20203dssd} proposes a hybrid sampling strategy by utilizing both feature and geometry distance for better classification performance.
Some approaches~\cite{chen2022sasa,wang2022rbgnet,zhang2022iassd} use objectness scores rather than feature distance to improve the foreground points ratio after downsampling.
It is generally time-consuming to repeatedly apply sampling and grouping operations on large-scale point clouds.

Point-voxel-based methods~\cite{ye2020hvnet,miao2021pvgnet,yang2022graphrcnn} take advantage of the efficiency of 3D sparse convolutions while preserving accurate point locations.
PV-RCNN~\cite{shi2020pvrcnn} and its variants~\cite{shi2021pvrcnnplusplus} extract point-wise features from voxel abstraction networks to refine the proposals generated from the 3D voxel backbone.
Pyramid R-CNN~\cite{mao2021pyramid} collects points for each RoI in a pyramid manner.

\textbf{3D Object Detection from Range Images.}
There are some prior works~\cite{alex2020rcd,sun2021rsn,liang2021rangeioudet} that attempt to predict 3D boxes from raw representations captured by LiDAR sensors, i.e., 2D perspective range images.
LaserNet~\cite{meyer2019lasernet} applies traditional 2D convolutions to range images to directly regress boxes.
RangeDet~\cite{fan2021rangedet} and PPC~\cite{chai2021ppc} introduce point-based convolution kernels to capture 3D geometric information from 2D range view representation.
The representation of range images is compact and dense, and free of quantization errors, which inspires us to use it to speed up the ball query algorithm~\cite{qi2017pointnet++} widely used in point-based methods.
The difference from earlier methods is that our constructed virtual range images can handle more complex situations, such as point clouds from multi-frame or multi-sensor.

\textbf{Point Cloud Analysis by Transformer.}
Transformer~\cite{ashish2017transformer} has demonstrated its great success in many computer vision tasks~\cite{carion2020detr,zhu2021deformabledetr}.
Recent approaches~\cite{lai2022stratified,park2022fastpointtransformer,yang20213dman,liu2021sparsepoint,zhou2022centerformer,liu20223dqueryis,zhao2021pointtransformer,pan2021pointformer,liu20223dqueryis} also explore it for point cloud analysis.
Pointformer~\cite{pan2021pointformer} proposes local and global attention modules to process 3D point clouds.
Group-Free~\cite{liu2021groupfree} eliminates hand-crafted grouping schemes by applying an attention module on all the points.
3DETR~\cite{misra20213detr} develops an end-to-end Transformer-based detection model with minimal 3D specific inductive biases.
VoTr~\cite{mao2021votr} introduces a voxel-based Transformer that adopts both local and dilated attention to enlarge receptive fields of the model.
SST~\cite{fan2022sst} extends the shifted window~\cite{liu2021swin} to 3D scenes and employs self-attention on non-empty voxels within the window.
Object DGCNN~\cite{wang2021objectdgcnn} incorporates grid-based BEV features around queries through deformable attention~\cite{zhu2021deformabledetr}.
VISTA~\cite{deng2022vista} adaptively fuses global multi-view features via an attention module.
Despite the effectiveness, they often fail to capture fine patterns of point clouds due to voxelization.
CT3D~\cite{sheng2021ct3d} builds Transformer on top of a two-stage detector and operates attention on the points grouped by RoIs. 
EQ-PVRCNN~\cite{yang2022eqpvrcnn} takes proposal grids as queries and generates RoI features from a voxel-based backbone.

\begin{figure*}[!t]
	\centering
	\includegraphics[width=1.6\columnwidth]{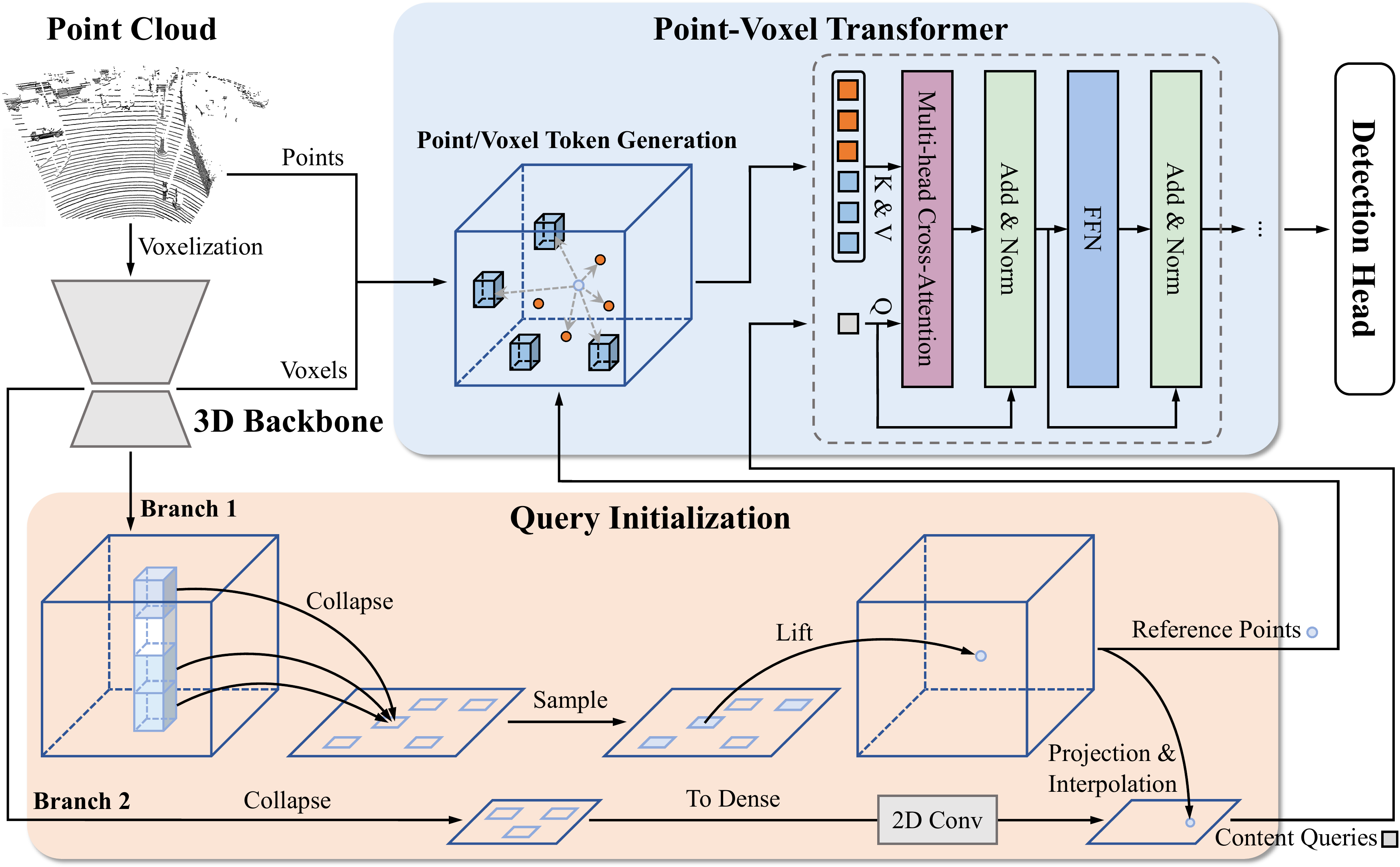}
	\vspace{-0.8em}
	\caption{Architecture.
	The raw point clouds are voxelized to feed into the sparse convolutions.
	\textbf{Query Initialization} module obtains reference points by collapsing, sampling, and lifting non-empty voxels, and then attains the corresponding content queries from the dense BEV feature map.
	They are processed by the \textbf{Point-Voxel Transformer} to adaptively fuse voxel and point features.
	Finally, the detection head uses fused features to produce classification and regression.}
	\label{fig:pvt_stru}
        \vspace{-1.4em}
\end{figure*}

\section{Methodology}

The architecture of our model is illustrated in Figure~\ref{fig:pvt_stru}.
Unlike previous Transformer-based indoor 3D detectors~\cite{misra20213detr,liu2021groupfree}, our target application is for outdoor scenes where point clouds are sparse and large-scale.
Thus, we use sparse convolutions as the backbone for efficient feature encoding.
Then the query initialization (in Sec.~\ref{method:query init}) obtains reference points from the non-empty voxels and content queries from the dense BEV feature map.
After that, the point-voxel Transformer (in Sec.~\ref{method:pv transformer}) samples voxels and points around reference points and utilizes their positions and features to capture contextual and geometric features via Transformer blocks.
To quickly find neighboring points near reference points, we propose the virtual range image (in Sec.~\ref{method:virtual range image}) to speed up the process.
Finally, we show the definition of the loss function (in Sec.~\ref{method:loss_func}).

\subsection{Query Initialization}
\label{method:query init}
Recent methods~\cite{liu2021groupfree,misra20213detr,yang2022eqpvrcnn} show that taking input-dependent queries (i.e., Q for the attention layer) benefits object detection.
For example, 3DETR~\cite{misra20213detr} for indoor 3D detection samples a set of reference points from inputs through FPS and associates each point with a content query.
However, it is not friendly for outdoor scenes because sampling from large-scale points is time-consuming.
To solve this problem, we propose a novel query initialization module that samples voxels instead of points to reduce the sampling time. 
Meanwhile, we make the voxels near objects more likely to be sampled, which can further boost the detection performance.
Specifically, the module has two branches: branch 1 samples 2D non-empty voxels and lifts the 2D voxels to generate 3D reference points; branch 2 first constructs the dense feature map through several convolutional layers and then indexes features in terms of branch 1's coordinates as content queries.

\textbf{Branch 1.}
For all input 3D voxels, we first collapse the voxels along the height dimension to further reduce the number of non-empty voxels.
To keep efficiency, we adopt the max pooling for voxels locating the same horizontal location.
After that, a representative set of voxels will be sampled.
Ideally, we would like the sampled voxels to recall as many of the foreground objects as possible.
Inspired by S-FPS~\cite{chen2022sasa}, we predict whether voxels are in objects and sample some of them according to their probability and geometric distance.
This allows voxels with better initial positions to be sampled and avoids sampled voxels being spatially too close.
Concretely, the predicted probability is supervised
by the foreground mask, i.e., if the voxel is inside a box, its label is set to 1; otherwise, it is 0.
We then multiply the foreground probability by the original voxel distance as the new distance metric, and iteratively sample the most distant voxel.
Finally, we use the center of objects as a guide to lift 2D voxels to 3D points.
That is, for each sampled voxel, it predicts the offsets to the center of its corresponding 3D boxes, and the height of voxels is set to 0 by default.
The predicted offsets are then added to the coordinates of voxels to acquire 3D reference points $\mathcal{P}_\mathrm{query} \subset \mathbb{R}^3$.

\textbf{Branch 2.}
As the generated reference points have deviated from the coordinates of the original voxels, direct use of the original voxel features as the corresponding content queries will result in a mismatch between instance features and positions.
To overcome this, we use voxel features from the middle layer of the 3D backbone to align the new positions by projection and interpolation.
Firstly, the 3D voxels are collapsed into the sparse BEV feature map, which is then converted to dense, i.e., empty voxels are filled with zeros.
Subsequently, several lightweight convolutional layers are applied for feature extraction to avoid the generated reference points indexing invalid features from empty voxels.
After that, we project reference points to 2D BEV coordinate system and index features on the BEV feature map as content queries $\mathcal{F}_\mathrm{query} \subset \mathbb{R}^d$.
Since the projected coordinates may not be integers, we use bilinear interpolation to collect the feature vector.

\subsection{Point-Voxel Transformer}
\label{method:pv transformer}
The point-voxel Transformer takes four inputs: reference points, content queries, raw points, and voxels from the 3D backbone.
The voxel tokens and point tokens are generated, which are then fed into several Transformer blocks to adaptively captures long-range contextual features and fine-grained geometric features.

\textbf{Voxel Token Generation.}
To capture long-range contexts, some methods~\cite{liu2021groupfree,pan2021pointformer,deng2022vista} perform attention on all points or sampled points, which are inefficient.
The former has too many points (e.g., 180K for point clouds and 20K for non-empty voxels), while the latter relies on FPS for downsampling.
Recent studies~\cite{misra20213detr} show that local feature aggregation matters more than global aggregation.
Motivated by that, to reduce computational and memory overheads while keeping large receptive fields, we randomly sample $l$ voxels from the middle layer of the 3D backbone within a large radius $r_v$ (i.e., 8m) of reference points as the voxel tokens, which consists of voxel coordinates $\mathcal{P}_\mathrm{voxel}$ and voxel features $\mathcal{F}_\mathrm{voxel}$.

\begin{figure}[!t]
	\centering
	\includegraphics[width=1.0\columnwidth]{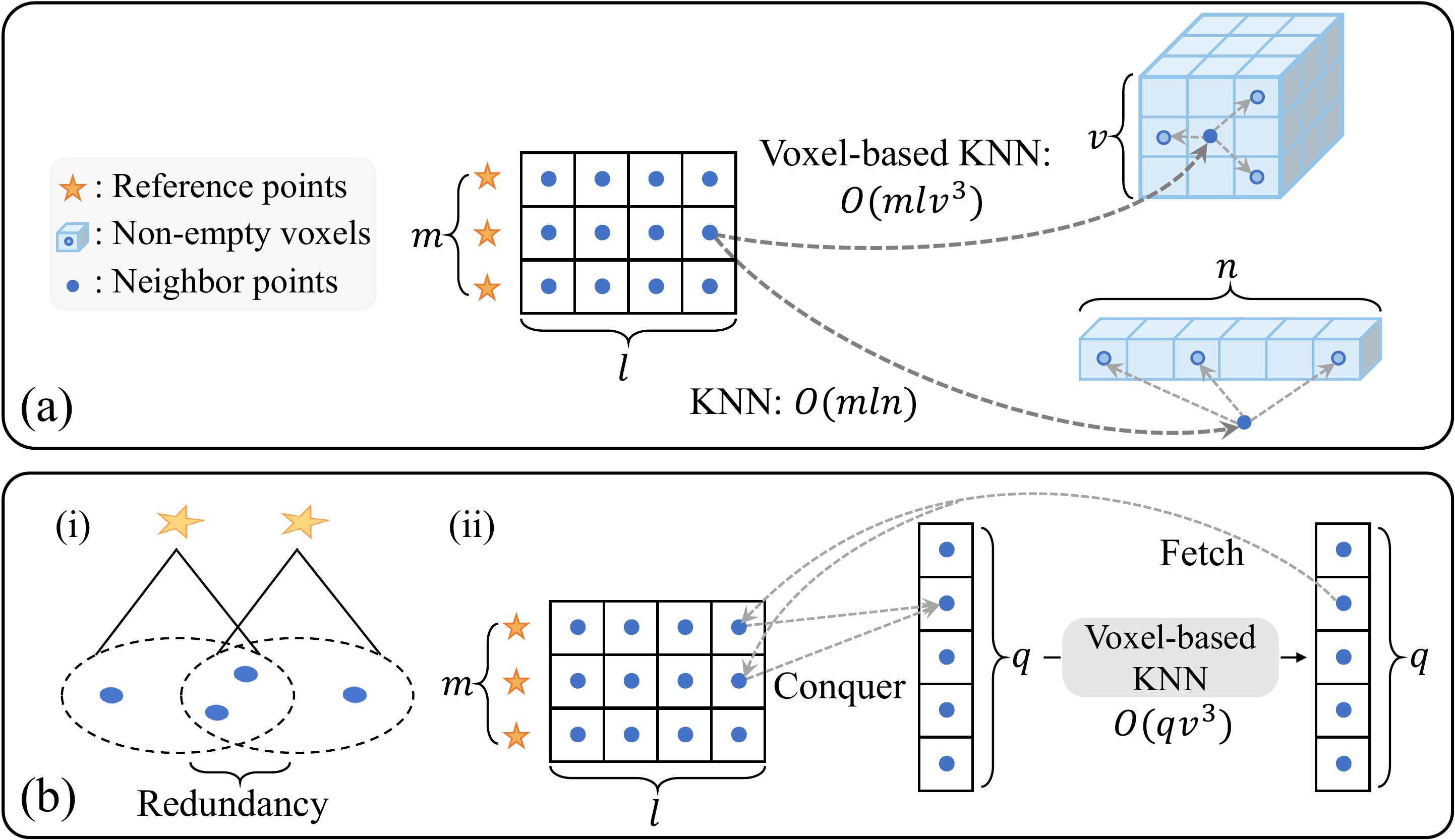}
	\vspace{-1.8em}
	\caption{Illustration of the voxel-based KNN (a) and the conquer-fetch operation (b).}
	\label{fig:interpolate}
        \vspace{-1.4em}
\end{figure}

\textbf{Point Token Generation.}
Due to the quantization artifacts produced during voxelization, a great hurdle remains for the voxel tokens in producing accurate 3D boxes.
Therefore, we generate the point tokens, which have smaller receptive fields than the voxel tokens but can provide fine-grained point features.
Specifically, we apply the ball query (introduced in Sec.~\ref{method:virtual range image}) to sample $l$ points within the radius $r_p$ (i.e., 3.2m) of reference points.
However, the sampled points $\mathcal{P}_\mathrm{point}$ only contain xyz position information, lacking local geometric and contextual information.
As a result, the attention map fails to capture the high-level correlation between the query and key in the Transformer block.
Inspired by \cite{he2020sassd,liu2019pvcnn}, we interpolate the features of 3D voxels near the points to obtain point features $\mathcal{F}_\mathrm{point}$:
\begin{equation}
\left\{f_i=\frac{\sum_{j=1}^k w_j^i \cdot \bar{f}_j^i}{\sum_{j=1}^k w_j^i}\ \bigg|\ w_j^i = \frac{1}{\left\|\bar{p}_j^i-p_i\right\|_2}\right\},
\end{equation}
where $p_i\in\mathcal{P}_\mathrm{point}$ and $f_i\in\mathcal{F}_\mathrm{point}$ are the coordinate and feature of the $i$-th sampled point, respectively, $\bar{p}_j^i$ and $\bar{f}_j^i$ are the center and feature of the $j$-th voxel near $p_i$, respectively, and $k$ is the number of neighbors.
We use k-nearest neighbors (KNN) to acquire $\bar{p}_j^i$ for each $p_i$.
However, the original KNN implemented by PyTorch costs $O(mln)$ to find neighbors, as shown in Figure~\ref{fig:interpolate}(a).
Inspired by \cite{deng2021voxelrcnn}, we propose the voxel-based KNN that only searches nearby voxels instead of all of them to reduce the complexity to $O(mlv^3)$.
To further diminish the computational costs, we design the conquer-fetch operation (in (ii) of Figure~\ref{fig:interpolate}(b)).
We empirically observe that some redundant points are sampled for different reference points (in (i) of Figure~\ref{fig:interpolate}(b)), but their voxel neighbors are the same.
Therefore, the conquer operation is first applied to remove redundancy, followed by the voxel-based KNN to find voxel neighbors. The fetch operation is then applied to obtain voxel neighbors of all points.
The time complexity is further decreased to $O(qv^3)$.

\textbf{Transformer Block.}
Given matrices of the voxel tokens $\mathcal{F}_\mathrm{voxel}$ and $\mathcal{P}_\mathrm{voxel}$, and the point tokens $\mathcal{F}_\mathrm{point}$ and $\mathcal{P}_\mathrm{point}$, we first concatenate voxel tokens and point tokens to construct $\mathcal{F}_s=[\mathcal{F}_\mathrm{voxel}, \mathcal{F}_\mathrm{point}]$ and $\mathcal{P}_s=[\mathcal{P}_\mathrm{voxel},\mathcal{P}_\mathrm{point}]$.
Combined with the content queries $\mathcal{F}_\mathrm{query}$ and the reference points $\mathcal{P}_\mathrm{query}$, they are then fed into a Transformer block:
\begin{equation}
	\begin{aligned}
	X &= \text{Attention}(\mathcal{F}_s, \mathcal{F}_\mathrm{query}, \mathcal{P}_s, \mathcal{P}_\mathrm{query}) + \mathcal{F}_\mathrm{query},\\
	Y &= \text{FFN}(X) + X,
	\end{aligned}
\end{equation}
where Attention is a multi-head cross-attention layer with contextual relative positional encoding~\cite{wu2021crpe,yang2022eqpvrcnn}, and FFN is a feed-forward network.
We use LayerNorm~\cite{ba2016ln} to normalize features after each Attention and FFN module.

\begin{figure*}[!t]
	\centering
	\includegraphics[width=1.6\columnwidth]{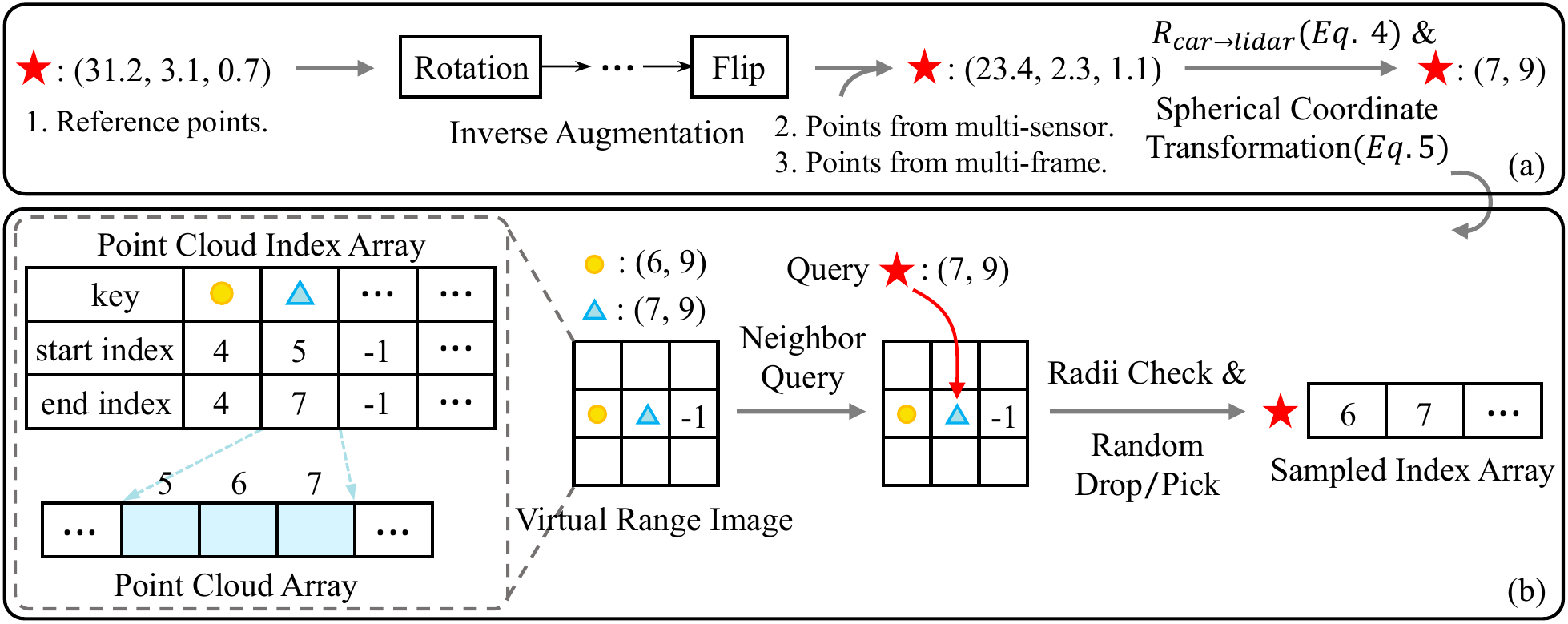}
	\vspace{-0.8em}
	\caption{Illustration of the proposed ball query, which consists of two parts: (a) Inverse augmentation and coordinate transformation; (b) We construct the virtual range image where the pixel value is represented as Point Cloud Index Array.
	The index array records the overlapped 3D points on the range image.
	Then we use Neighbor Query to find the neighbors of reference points, and apply Radii Check and Random Drop/Pick to sample 3D points.}
	\label{fig:rv_rand}
	\vspace{-1.4em}
\end{figure*}

\subsection{Virtual Range Image}
\label{method:virtual range image}
As mentioned in point token generation (in Sec.~\ref{method:pv transformer}), we introduce a novel ball query to quickly find neighbors of reference points.
As illustrated in Figure~\ref{fig:rv_rand}, our ball query is based on the virtual range image that is constructed from the point clouds of multi-sensor and multi-frame.

\textbf{Setup.}
Let $\mathcal{R}_{\mathrm{lidar} \rightarrow \mathrm{car}}^i \in \mathbb{R}^{4 \times 4}$ be a homogeneous transformation matrix that transforms points $\mathcal{P}_\mathrm{lidar}^i \subset \mathbb{R}^3$ of each LiDAR sensor $\mathcal{S}_i$ from the sensor coordinate system to the car coordinate system.
As a common practice, we use all points $\mathcal{P}_\mathrm{car}$ from five LiDAR sensors on the Waymo dataset:
\begin{equation}
	\mathcal{P}_\mathrm{car}=\left\{
	\mathcal{P}_\mathrm{lidar}^i \cdot \mathcal{R}_{\mathrm{lidar} \rightarrow \mathrm{car}}^i\ |\ i \in \left\{0,\ldots,4\right\}
	\right\}.
\end{equation}

\textbf{Inverse Augmentation.}
In 3D object detection, augmentation (e.g., copy-n-paste~\cite{yan2018second}, global rotation, and random flip) plays a vital role in reducing model overfitting.
For copy-n-paste, the object points $\mathcal{P}_\mathrm{gt}$ from other frames are pasted on the current frame with their original 3D positions.
We combine $\mathcal{P}_\mathrm{gt}$ and $\mathcal{P}_\mathrm{car}$ to get a new point set $\mathcal{P}$, which will be used to construct the virtual range image.
For other geometry-related data augmentation, we save augmentation parameters (e.g., the rotation angle for global rotation).
Since the reference points $\mathcal{P}_\text{query}$ (we use $\mathcal{Q}$ to simplify the notation) are generated after augmentation, we reverse all those data augmentations on $\mathcal{Q}$ to get the original coordinate (in Figure~\ref{fig:rv_rand}(a)), which is similar to \cite{zhang2020moca,li2022deepfusion}.

\textbf{Coordinate Transformation.}
Next, we transform $\mathcal{P}$ and $\mathcal{Q}$ from the reference frame of the car back to the top LiDAR sensor (i.e., $\mathcal{S}_0$):
\begin{equation}
\mathcal{P}^\prime = \mathcal{P} \cdot \mathcal{R}_{\mathrm{car} \rightarrow \mathrm{lidar}}^0,
\quad \mathcal{Q}^\prime = \mathcal{Q} \cdot \mathcal{R}_{\mathrm{car} \rightarrow \mathrm{lidar}}^0,
\end{equation}
where $\mathcal{P}^\prime$ and $\mathcal{Q}^\prime$ are the coordinates in the LiDAR Cartesian coordinate system.
For each point $\boldsymbol{p}=(x, y, z) \in \mathcal{P}^\prime \cup \mathcal{Q}^\prime$, it is uniquely transformed to the LiDAR Spherical coordinate system by the following equations:
\begin{equation}
	\begin{aligned}
		\theta&=\atantwo (y, x),\quad \phi=\atantwo (z, \sqrt{x^2 + y^2}),\\
		r&=\sqrt{x^2 + y^2 + z^2},
	\end{aligned}
\end{equation}
where $\theta$, $\phi$, and $r$ indicate the laser's azimuth angle, inclination angle, and range, respectively.

\begin{figure}[!t]
	\centering
        \vspace{0.2em}
	\includegraphics[width=1.0\columnwidth]{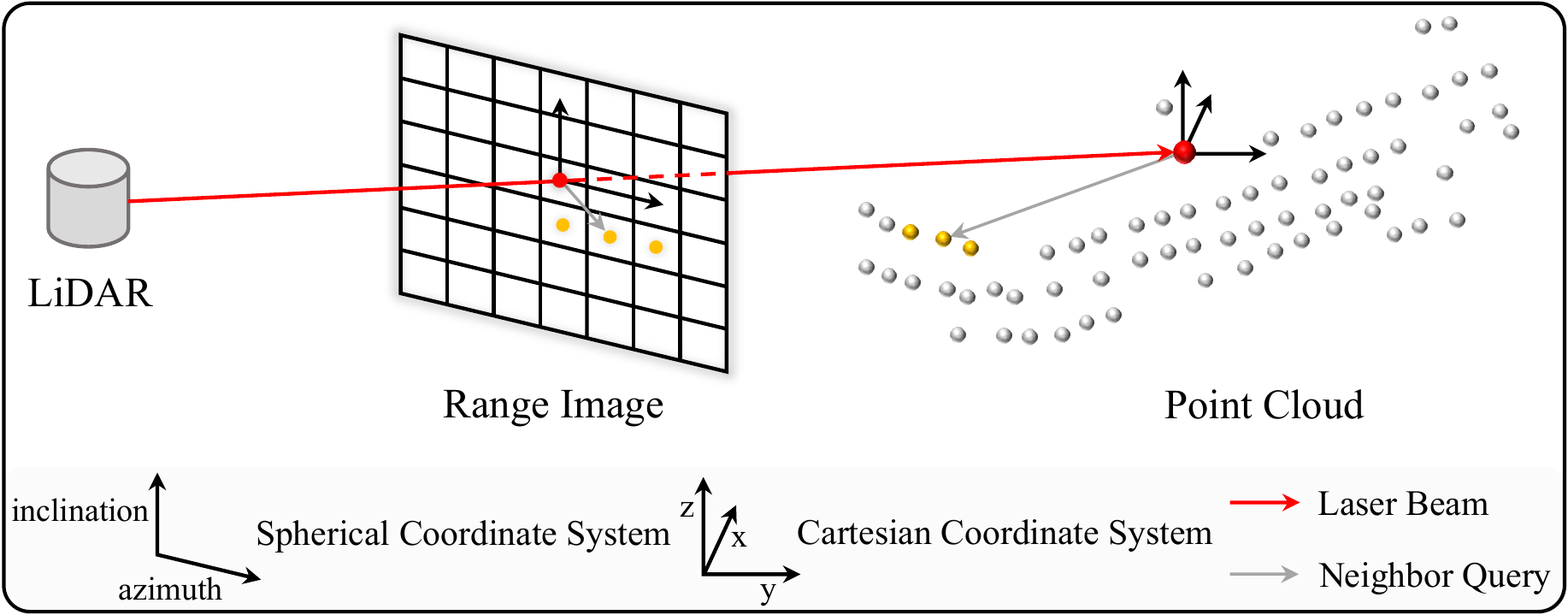}
	\vspace{-1.8em}
	\caption{Illustration of the Cartesian and the Spherical coordinate system of the LiDAR sensor.}
	\label{fig:lidar_demo}
	\vspace{-1.5em}
\end{figure}

\textbf{Image Construction.}
Suppose $\bar{\mathcal{P}}$ and $\bar{\mathcal{Q}}$ are transformed coordinates of $\mathcal{P}^\prime$ and $\mathcal{Q}^\prime$, respectively.
The native range image uses $\theta$ as the column index, $\phi$ as the row index, and $r$ as the pixel value, which has a one-to-one correspondence with the point cloud captured by one LiDAR sensor, as shown in Figure~\ref{fig:lidar_demo}.
However, as $\bar{\mathcal{P}}$ is from multiple frames as well as multiple LiDAR sensors, various points may overlap on the same pixel.
To overcome this issue, each pixel of our constructed virtual range image is an array of indices pointing to overlapped 3D points.
For convenience, we organize overlapped points adjacently and record their start and end position indices, as shown by Point Cloud Index Array and Point Cloud Array in Figure~\ref{fig:rv_rand}(b).

\textbf{Random Neighbor Query.}
With the constructed image, each reference point in $\bar{\mathcal{Q}}$ can quickly find its neighbors because $\theta$ and $\phi$ determine the pixel position in the virtual range image, as shown by the Neighbor Query in Figure~\ref{fig:rv_rand}(b).
To prevent the distance between adjacent pixels from being large in 3D space, we will check whether the neighbors are within the radius of the reference point (i.e., Radii Check).
During the process, we will apply a random selection algorithm\footnote{\url{https://github.com/LeviViana/torch_sampling}} to retain at most $k$ neighbors (i.e., Random Drop/Pick).
The overall operation can be efficiently executed in parallel on the GPU.
In this way, we reduce the theoretical time complexity of the original ball query~\cite{qi2017pointnet++} from $\mathcal{O}(mn)$ to $\mathcal{O}(ms^2)$, where $m$ is the number of reference points, $n$ is the number of point clouds, and $s$ is the kernel size to visit neighbors in the virtual range image. Notably, $s^2$ is much smaller than $n$.

\subsection{Loss Function}
\label{method:loss_func}
The overall loss consists of the segmentation loss, offset loss, classification loss, and regression loss.

\textbf{Segmentation Loss \& Offset Loss.}
We regard non-empty voxels inside any ground-truth bounding boxes as foreground voxels.
Since voxels are two-dimensional, the height of the ground-truth box is not considered when assigning labels. The segmentation loss is computed by:
\begin{equation}
	\mathcal{L}_\mathrm{seg}=\frac{1}{N^v} \cdot \sum_{i=1}^{N^v} \text{CE}\left(s_i, \hat{s}_i\right),
\end{equation}
where $N^v$ is the number of non-empty voxels, and CE is the cross entropy loss function.
The offset loss is calculated by:
\begin{equation}
	\mathcal{L}_\mathrm{offset}=\frac{1}{N^v_+} \cdot \sum_{i=1}^{N^v_+} \text{L1}\left(o_i, \hat{o}_i\right),
\end{equation}
where $N^v_+$ is the number of foreground voxels, and L1 is the smooth-$l_1$ loss function.

\textbf{Classification Loss \& Regression Loss.}
We adopt the same target assignment strategy and prediction head following \cite{yang20203dssd,chen2022sasa,zhang2022iassd}.
Specifically, for each reference point, we consider the point inside an annotated bounding box as the foreground point and then compute the centerness~\cite{yang20203dssd} as its label.
The classification loss is:
\begin{equation}
	\mathcal{L}_\mathrm{cls}=\frac{1}{N^q} \cdot \sum_{i=1}^{N^q} \text{CE}\left(c_i, \hat{c}_i\right),
\end{equation}
where $N^q$ is the number of reference points.
For the regression loss $\mathcal{L}_\mathrm{reg}$, we decouple it to center regression loss, box size estimation loss, and heading angle estimation loss.
We refer readers to \cite{yang20203dssd} for more details.
\section{Experiments}
\subsection{Datasets}
\textbf{Waymo Open Dataset}~\cite{sun2020wod} is a large-scale autonomous driving dataset consisting of 798 scenes for training and 202 scenes for validation.
The evaluation protocol consists of the average precision (AP) and average precision weighted by heading (APH).
Also, it includes two difficulty levels: LEVEL\_1 denotes objects containing more than 5 points, and LEVEL\_2 denotes objects containing at least 1 point.
To save training time, we use a subset of the training splits by sampling every 10 frames for ablation studies.

\textbf{KITTI}~\cite{geiger2012kitti} contains 7481 training samples and 7518 testing samples in autonomous driving scenes.
As a common practice, the training data are divided into a \emph{train} set with 3712 samples and a \emph{val} set with 3769 samples.

\textbf{nuScenes}~\cite{caesar2020nuscenes} is a challenging dataset for autonomous driving with 380K LiDAR sweeps from 1000 scenes.
The evaluation metrics used in nuScenes dataset incorporate the commonly used mean average precision (mAP) and a novel nuScenes detection score (NDS).

\begin{table}[t]
	\centering
	\caption{Performance comparison on the Waymo validation set for vehicle class. 3f: taking 3 frames as input.  The results achieved by our PVT-SSD are shown in bold, while the top-performed results are shown in underline.}
	\vspace{-1.0em}
	\resizebox{0.74\columnwidth}!{\begin{tabular}{l|cc}
			\toprule
			\multirow{2}{*}{Methods} & \multirowcell{2}{LEVEL\_1\\3D AP/APH} & \multirowcell{2}{LEVEL\_2\\3D AP/APH} \\
			& & \\
			\midrule
			\bf{Two-stage:} & & \\
			RSN~\cite{sun2021rsn} & 75.10/74.60 & 66.00/65.50 \\
			Pyramid RCNN~\cite{mao2021pyramid} & 76.30/75.68 & 67.23/66.68 \\
			SST\_TS~\cite{fan2022sst} & 76.22/75.79 & 68.04/67.64 \\
			LiDAR R-CNN~\cite{li2021lidarrcnn} & 76.00/75.50 & 68.30/67.90 \\
			Part-A2-Net~\cite{shi2020part} & 77.05/76.51 & 68.47/67.97 \\
			CenterPoint-Voxel~\cite{yin2021center} & 76.70/76.20 & 68.80/68.30 \\
			PV-RCNN~\cite{shi2020pvrcnn} & 77.51/76.89 & 68.98/68.41 \\
			CT3D~\cite{sheng2021ct3d} & 76.30/- & 69.04/- \\
			PDV~\cite{hu2022pdv} & 76.85/76.33 & 69.30/68.81 \\
			BtcDet~\cite{xu2022btcdet} & 78.58/78.06 & 70.10/69.61 \\
			PV-RCNN++~\cite{shi2021pvrcnnplusplus} & 79.25/78.78 & 70.61/70.18 \\
			\midrule
			\bf{One-stage:} & & \\
			IA-SSD~\cite{zhang2022iassd} & 70.53/69.67 & 61.55/60.80 \\
			PointPillars~\cite{lang2019pointpillar} & 71.56/70.99 & 63.05/62.54 \\
			SECOND~\cite{yan2018second} & 72.27/71.69 & 63.85/63.33 \\
			RangeDet~\cite{fan2021rangedet} & 72.90/72.30 & 64.00/63.60 \\
			CenterPoint-Pillar~\cite{yin2021center} & 73.37/72.86 & 65.09/64.62 \\
                SST~\cite{fan2022sst} & 74.22/73.77 & 65.47/65.07 \\
			VoxSeT~\cite{he2022voxset} & 74.50/74.03 & 65.99/65.56 \\
			CenterPoint-Voxel~\cite{yin2021center} & 74.78/74.24 & 66.66/66.17 \\
			Point2Seq~\cite{xue2022point2seq} & 77.52/77.03 & 68.80/68.36 \\
                MsSVT~\cite{dong2022mssvt} & 77.83/77.32 & 69.53/69.06 \\
			SWFormer~\cite{sun2022swformer} & 77.80/77.30 & 69.20/68.80 \\
                SWFormer\_3f~\cite{sun2022swformer} & 79.40/78.90 & 71.10/70.60 \\
			\midrule
            \textbf{PVT-SSD (Ours)} & \textbf{79.16}/\textbf{78.72} & \textbf{70.23}/\textbf{69.83} \\
            \textbf{PVT-SSD\_3f (Ours)} & \underline{\textbf{80.59}}/\underline{\textbf{80.16}} & \underline{\textbf{71.86}}/\underline{\textbf{71.47}} \\ 
			\bottomrule 
	\end{tabular}}
	\label{table:waymo_vehicle_val}
	\vspace{-1.0em}
\end{table}

\begin{table}[!t]
	\centering
	\caption{Performance comparison on the Waymo validation set for pedestrian and cyclist classes.}
	\vspace{-1.0em}
	\resizebox{0.74\columnwidth}!{\begin{tabular}{l|cc}
			\toprule
			Methods & Pedestrian & Cyclist \\
			\midrule
			\bf{Two-stage:} & & \\
			LiDAR R-CNN~\cite{li2021lidarrcnn} & 63.10/51.70 & 66.10/64.40 \\
			PV-RCNN~\cite{shi2020pvrcnn} & 66.04/57.61 & 65.39/63.98 \\
			PDV~\cite{hu2022pdv} & 65.85/58.28 & 66.49/65.36 \\
			Part-A2-Net~\cite{shi2020part} & 66.18/58.62 & 66.13/64.93 \\
			RSN~\cite{sun2021rsn} & 68.30/63.70 & -/- \\
			PV-RCNN++~\cite{shi2021pvrcnnplusplus} & 73.17/68.00 & 71.21/70.19 \\
			\midrule
			\bf{One-stage:} & & \\
                SST~\cite{fan2022sst} & 70.02/61.67 & -/- \\
                CenterPoint-Voxel~\cite{yin2021center} & 68.42/62.67 & 69.69/68.59 \\
			RangeDet~\cite{fan2021rangedet} & 67.60/63.90 & 63.30/62.10 \\
			VoxSeT~\cite{he2022voxset} & 72.45/65.39 & 68.95/67.73 \\
                MsSVT~\cite{dong2022mssvt} & 73.00/66.65 & 72.37/71.24 \\
                SWFormer~\cite{sun2022swformer} & 72.50/64.90 & -/- \\
                SWFormer\_3f~\cite{sun2022swformer} & 74.80/71.10 & -/- \\
			\midrule
            \textbf{PVT-SSD (Ours)} & \textbf{72.56}/\textbf{67.02} & \textbf{73.94}/\textbf{72.96} \\ 
            \textbf{PVT-SSD\_3f (Ours)} & \underline{\textbf{75.11}}/\underline{\textbf{72.12}} & \underline{\textbf{74.80}}/\underline{\textbf{73.97}} \\ 
			\bottomrule 
	\end{tabular}}
	\label{table:waymo_other_val}
	\vspace{-1.4em}
\end{table}

\subsection{Implementation Details}
Our implementation is based on the codebase of OpenPCDet\footnote{\url{https://github.com/open-mmlab/OpenPCDet}}.
For the Waymo dataset, the detection ranges are set as $(-75.2, 75.2)$, $(-75.2, 75.2)$, and $(-2, 4)$, and the voxel size is $(0.1m, 0.1m, 0.15m)$.
We adopt a similar 3D sparse backbone as \cite{ye2022lidarmultinet}, but the last two upsampling layers are removed to keep efficiency.
For the query initialization, we apply $4$ traditional 2D convolutional layers with dimensions $64$ on the BEV feature map; we sample $512$ voxels to generate reference points.
For the point-voxel Transformer, we sample $128$ neighbors for each reference point and apply one Transformer block; the $r_v$ and $r_p$ are set to $8.0$ and $3.2$, respectively; for each Transformer block, the input dimension, the hidden dimension, the number of head, and the dropout are set to $128$, $512$, $4$, and $0.1$, respectively; for the point token generation, each point interpolates from voxel features of the $8$ nearest neighbors.
We train the model for $30$ epochs with the AdamW~\cite{losh2019adamw} optimizer using the one-cycle policy, with a max learning rate of $3e^{-3}$.

\begin{table}[!t]
	\centering
        \setlength\tabcolsep{3pt}
	\caption{Performance comparison on the Waymo leaderboard.}
	\vspace{-1.0em}
	\resizebox{1.0\columnwidth}!{\begin{tabular}{l|c|ccc}
			\toprule
			Methods & mAP/mAPH & Vehicle & Pedestrian & Cyclist \\
			\midrule
                PV-RCNN~\cite{shi2020pvrcnn} & 71.25/68.75 & 72.81/72.39 & 71.81/66.05 & 69.13/67.80 \\
                PV-RCNN++~\cite{shi2021pvrcnnplusplus} & 72.42/70.20 & 73.86/73.47 & 74.12/69.00 & 69.28/68.15 \\
                Graph-RCNN~\cite{yang2022graphrcnn} & 73.81/71.59 & \underline{76.04}/\underline{75.64} & 75.59/70.45 & 69.79/68.67 \\
                GD-MAE~\cite{yang2023gd-mae} & \underline{74.71}/72.29 & 75.83/75.46 & \underline{77.10}/71.28 & 71.21/70.15 \\
                FSD~\cite{fan2022fsd} & 74.39/72.35 & 74.40/74.06 & 75.93/71.26 & 72.85/71.75 \\
                CenterPoint\_2f~\cite{yin2021center} & 73.38/71.93 & 73.42/72.99 & 74.56/71.52 & 72.17/71.28 \\
			SST\_TS\_3f~\cite{fan2022sst} & 74.41/72.81 & 73.08/72.74 & 76.93/\underline{73.51} & \underline{73.22}/72.17 \\
                SWFormer\_3f~\cite{sun2022swformer} & -/- & 75.02/74.65 & 75.87/72.07 & -/- \\
			\midrule
            \textbf{PVT-SSD (Ours)} & \textbf{72.73}/\textbf{70.44} & \textbf{72.96}/\textbf{72.62} & \textbf{73.15}/\textbf{67.72} & \textbf{72.08}/\textbf{70.97} \\ 
            \textbf{PVT-SSD\_3f (Ours)} & \textbf{74.69}/\underline{\textbf{73.28}} &\textbf{75.24}/\textbf{74.89} & \textbf{75.62}/\textbf{72.60} & \textbf{73.21}/\underline{\textbf{72.35}} \\ 
			\bottomrule 
	\end{tabular}}
	\label{table:waymo_test}
	\vspace{-1.2em}
\end{table}

\subsection{Comparison with State-of-the-Art Methods}
We compare PVT-SSD on the Waymo validation set with previous methods in Table~\ref{table:waymo_vehicle_val} and Table~\ref{table:waymo_other_val}.
We divide current methods into the branches of one-stage and two-stage detectors for comparisons.
Table~\ref{table:waymo_vehicle_val} shows the results on vehicles, where PVT-SSD surpasses previous one-stage methods with a single frame LiDAR input.
The performance is also comparable with state-of-the-art two-stage methods.
Table~\ref{table:waymo_other_val} shows results on pedestrians and cyclists. 
PVT-SSD outperforms the strong single-stage detector, i.e., CenterPoint-Voxel, by 4.35 APH and 4.37 APH for the pedestrian and cyclist, respectively.
The performance can be further improved by 5.1 APH and 1.01 APH, respectively, when taking 3 frames as input.
Table~\ref{table:waymo_test} illustrates the performance on the Waymo test set.
We achieve competitive results compared with previous methods.
Table~\ref{table:kitti_test} illustrates the performance comparisons on the official KITTI test server.
It shows that PVT-SSD has the best car detection performance among all single-stage detectors at both easy and moderate levels.
Compared with previous Transformer-based detectors, PVT-SSD is 1.4$\times$ and 4.4$\times$ faster in inference speed than CT3D and VoTr-TSD, respectively, and achieves better moderate AP.
It also requires less number of parameters, as shown in Table~\ref{table:kitti_para}.
In Table~\ref{table:nuscenes_val}, we report the results on nuScenes validation set.
Our method obtains much higher NDS and mAP compared with SASA~\cite{chen2022sasa}.

\begin{table}[!t]
	\centering
    \setlength{\tabcolsep}{2pt}
	\caption{Performance comparison on the KITTI testing sever for the car class. $\dag$: the latency is measured by us based on the same environment (i.e., RTX 3080Ti GPU) with 1 batch size.}
	\vspace{-1.0em}
	\resizebox{1.0\columnwidth}!{\begin{tabular}{l|ccc|ccc|c}
			\toprule
			\multirow{2}{*}{Methods} & \multicolumn{3}{c|}{3D AP} & \multicolumn{3}{c|}{BEV AP} & \multirowcell{2}{Latency\\(ms)} \\
			& Easy & Moderate & Hard & Easy & Moderate & Hard \\
			\midrule
			\bf{Two-stage:} & & & & & & \\
			PointRCNN~\cite{shi2019pointrcnn} & 86.96 & 75.64 & 70.70 & 92.13 & 87.39 & 82.72 & 100 \\
			STD~\cite{yang2019std} & 87.95 & 79.71 & 75.09 & 94.74 & 89.19 & 86.42 & 80 \\
			PV-RCNN~\cite{shi2020pvrcnn} & 90.25 & 81.43 & 76.82 & 94.98 & 90.65 & 86.14 & 98$^\dag$ \\
			M3DETR~\cite{guan2022m3detr} & 90.28 & 81.73 & 76.96 & 94.41 & 90.37 & 85.98 & - \\
			CT3D~\cite{sheng2021ct3d} & 87.83 & 81.77 & 77.16 & 92.36 & 88.83 & 84.07 & 70$^\dag$ \\
            PDV~\cite{hu2022pdv} & 90.43 & 81.86 & 77.36 & 94.56 & 90.48 & 86.23 & 135 \\
			PV-RCNN++~\cite{shi2021pvrcnnplusplus} & 90.14 & 81.88 & 77.15 & 92.66 & 88.74 & 85.97 & 60 \\
            EQ-PVRCNN~\cite{yang2022eqpvrcnn} & 90.13 & 82.01 & 77.53 & 94.55 & 89.09 & 86.42 & 200 \\
			Pyramid-PV~\cite{mao2021pyramid} & 88.39 & 82.08 & 77.49 & 92.19 & 88.84 & 86.21 & 127 \\
			VoTr-TSD~\cite{mao2021votr} & 89.90 & 82.09 & \underline{79.14} & 94.03 & 90.34 & 86.14 & 216$^\dag$ \\
            SPG~\cite{xv2021spg} & 90.50 & 82.13 & 78.90 & 94.33 & 88.70 & 85.98 & 156 \\
			\midrule
			\bf{One-stage:} & & & & & & \\
			SECOND~\cite{yan2018second} & 83.34 & 72.55 & 65.82 & 89.39 & 83.77 & 78.59 & 50 \\
            PointPillars~\cite{lang2019pointpillar} & 82.58 & 74.31 & 68.99 & 90.07 & 86.56 & 82.81 & \underline{24} \\
			TANet~\cite{liu2019tanet} & 84.39 & 75.94 & 68.82 & 91.58 & 86.54 & 81.19 & 35 \\
			Point-GNN~\cite{shi2020pointgnn} & 88.33 & 79.47 & 72.29 & 93.11 & 89.17 & 83.90 & 643 \\
			3DSSD~\cite{yang20203dssd} & 88.36 & 79.57 & 74.55 & 92.66 & 89.02 & 85.86 & 38 \\ 
			SA-SSD~\cite{he2020sassd} & 88.75 & 79.79 & 74.16 & 95.03 & 91.03 & 85.96 & 40 \\
			IA-SSD~\cite{zhang2022iassd} & 88.87 & 80.32 & 75.10 & 93.14 & 89.48 & 84.42 & 44$^\dag$ \\
			SVGA-Net~\cite{he2022svganet} & 87.33 & 80.47 & 75.91 & 92.07 & 89.88 & 85.59 & 62 \\
			PVGNet~\cite{miao2021pvgnet} & 89.94 & 81.81 & 77.09 & 94.36 & 91.26 & \underline{86.63} & - \\
			SASA~\cite{chen2022sasa} & 88.76 & 82.16 & 77.16 & 92.87 & 89.51 & 86.35 & 63$^\dag$ \\
			\midrule
			\textbf{PVT-SSD (Ours)} & \underline{\textbf{90.65}} & \underline{\textbf{82.29}} & \textbf{76.85} & \underline{\textbf{95.23}} & \underline{\textbf{91.63}} & \textbf{86.43} & \textbf{49}$^\dag$ \\ 
			\bottomrule 
	\end{tabular}}
	\label{table:kitti_test}
	\vspace{-1.0em}
\end{table}

\begin{table}[!h]
	\centering
	\caption{Comparisons of the number of parameters.}
	\vspace{-1.0em}
	\resizebox{0.95\columnwidth}!{\begin{tabular}{l|ccccc}
			\toprule
			Methods & CT3D~\cite{sheng2021ct3d} & VoTr-TSD~\cite{mao2021votr} & M3DETR~\cite{guan2022m3detr} &  \textbf{PVT-SSD} \\
                \midrule
                \# Param. & 30M & 49M & 76M & \textbf{16M} \\
			\bottomrule 
	\end{tabular}}
	\label{table:kitti_para}
	\vspace{-1.4em}
\end{table}

\begin{table*}[!t]
    \centering
	\caption{Performance comparison on the nuScenes validation set. $\dag$: we follow \cite{yang20203dssd,chen2022sasa} to predict all classes in a single head.}
	\vspace{-1.0em}
	\resizebox{1.65\columnwidth}!{\begin{tabular}{l|cc|cccccccccc}
			\toprule
			Methods & NDS & mAP & Car & Truck & Bus & Trailer & C. V. & Ped. & Motor & Bicycle & T. C. & Barrier \\
			\midrule
            SECOND~\cite{yan2018second} & - & 27.1 & 75.5 & 21.9 & 29.0 & 13.0 & 0.4 & 59.9 & 16.9 & 0 & 22.5 & 32.2 \\
			PointPillars~\cite{lang2019pointpillar} & 44.9 & 29.5 & 70.5 & 25.0 & 34.4 & 20.0 & 4.5 & 59.9 & 16.7 & 1.6 & 29.6 & 33.2 \\
			3DSSD~\cite{yang20203dssd} & 56.4 & 42.7 & \underline{81.2} & \underline{47.2} & 61.4 & 30.5 & 12.6 & 70.2 & 36.0 & 8.6 & 31.1 & 47.9 \\
			SASA~\cite{chen2022sasa} & 61.0 & 45.0 & 76.8 & 45.0 & \underline{66.2} & \underline{36.5} & 16.1 & 69.1 & 39.6 & 16.9 & 29.9 & 53.6 \\
			\midrule
            \textbf{PVT-SSD$^\dag$ (Ours)} & \underline{\textbf{65.0}} & \underline{\textbf{53.6}} & \textbf{79.4} & \textbf{43.8} & \textbf{62.1} & \textbf{34.2} & \underline{\textbf{21.7}} & \underline{\textbf{79.8}} & \underline{\textbf{53.4}} & \underline{\textbf{38.2}} & \underline{\textbf{56.6}} & \underline{\textbf{67.1}} \\
			\bottomrule 
	\end{tabular}}
	\label{table:nuscenes_val}
	\vspace{-1.6em}
\end{table*}

\subsection{Ablation Study}
\label{experiment:ablation}
In this section, we conduct a series of ablation experiments to comprehend the roles of different components.

\begin{table}[!t]
	\centering
	\caption{Ablations on the Waymo validation set for vehicle class.}
	\vspace{-1.0em}
	\resizebox{0.76\columnwidth}!{\begin{tabular}{cccc|c}
			\toprule
			Lift. & Alig. & Voxel Tok. & Point Tok. & 3D AP/APH \\
			\midrule
			 & & & & 62.82/62.37 \\
			\checkmark & \checkmark & & & 63.40/62.98 \\
			\checkmark & \checkmark & \checkmark & & 63.85/63.43 \\
			\checkmark & & \checkmark & \checkmark & 64.57/64.14 \\
                \checkmark & \checkmark & & \checkmark & 65.12/64.70 \\
			\checkmark & \checkmark & \checkmark & \checkmark & \textbf{65.45}/\textbf{65.01} \\
			\bottomrule 
	\end{tabular}}
	\label{table:ablation_study}
        \vspace{-0.8em}
\end{table}

\textbf{Query Initialization.}
Lifting plays an essential role in query initialization.
It can not only make 2D voxels into 3D points, but also can bring reference points closer to the center of the object, which can make it easier for subsequent regressions to recall their corresponding 3D boxes.
As shown in the first and second rows of Table~\ref{table:ablation_study}, in combination with feature alignment, it brings 0.61 APH gains.
In Table~\ref{table:dif_sample}, we compare performances when using different sampling strategies.
Compared with FPS and F-FPS~\cite{yang20203dssd}, S-FPS yields 11.25 APH and 0.48 APH benefits, respectively.
It considers the probability of whether voxels are in objects when sampling, which allows it to sample voxels with better initial positions to generate reference points.
S-FPS is also 0.81 APH higher than Top-K because it avoids sampling voxels too closely in space, thus recalling foreground objects as much as possible.
The fourth and sixth rows of Table~\ref{table:ablation_study} show that feature alignment leads to a 0.87 APH improvement because the semantic similarity with voxel tokens and point tokens can be calculated more correctly.

\textbf{Point-Voxel Transformer.}
Table~\ref{table:ablation_study} shows that the voxel tokens and the point tokens provide improvements of 0.45 and 1.58 APH, respectively.
We find that the improvement brought by the point tokens is larger than that of the voxel tokens, which is intuitive since fine-grained point features are essential for accurate box regression.
Table~\ref{table:ablate_radii} ablates the influence of $r_v$ and $r_p$, which will affect the receptive field.
We observe that large receptive fields improve detection performance, but it is not further improved when continuing to increase the receptive field.
It may be due to the fact that more irrelevant noise points are sampled while relevant points benefiting detection are randomly discarded.
Table~\ref{table:pos_encoding} shows the importance of contextual relative position encoding, which has a huge impact on the learned representation.
When absolute position encoding~\cite{misra20213detr} and bias-mode relative position encoding~\cite{wu2021crpe} are used, the APH drops by 1.25 and 0.73, respectively.

\begin{table}[!t]
	\centering
	\caption{Ablations of sampling strategies in query initialization.}
	\vspace{-1.0em}
	\resizebox{0.65\columnwidth}!{\begin{tabular}{c|cccc}
			\toprule
			Methods & S-FPS & F-FPS & FPS & Top-K \\
			\midrule
			APH & \textbf{65.01} & 64.53 & 53.76 & 64.20 \\
			\bottomrule
	\end{tabular}}
	\label{table:dif_sample}
        \vspace{-0.8em}
\end{table}

\begin{table}[!t]
	\centering
	\caption{Ablations of radii $r_v$ and $r_p$ in point-voxel Transformer.}
	\vspace{-1.0em}
	\begin{subtable}[b]{0.48\columnwidth}
	\centering
	\resizebox{0.95\columnwidth}!{\begin{tabular}{c|ccc}
		\toprule
		$r_v$ & 6.4 & 8.0 & 9.6\\
		\midrule
		APH & 64.79 & \textbf{65.01} & 64.94 \\
		\bottomrule
	\end{tabular}}
	\end{subtable}
	\hfill
	\begin{subtable}[b]{0.48\columnwidth}
	\centering
	\resizebox{0.95\columnwidth}!{\begin{tabular}{c|ccc}
		\toprule
		$r_p$ & 1.6 & 3.2 & 4.8 \\
		\midrule
		APH & 64.21 & \textbf{65.01} & 64.97 \\
		\bottomrule
	\end{tabular}}
	\end{subtable}
	\label{table:ablate_radii}
	\vspace{-1.6em}
\end{table}

\begin{table}[!b]
	\centering
	\vspace{-1.6em}
	\caption{Ablations of positional encoding in Transformer blocks.}
	\vspace{-1.0em}
	\resizebox{0.7\columnwidth}!{\begin{tabular}{c|cccc}
		\toprule
		\multirow{2}{*}{Methods} & \multirow{2}{*}{None} & \multicolumn{2}{c}{Relative} & \multirow{2}{*}{Absolute} \\
		& & Context & Bias & \\
		\midrule
		APH & 63.29 & \textbf{65.01} & 64.28 & 63.76 \\
		\bottomrule
	\end{tabular}}
        \vspace{-1.0em}
	\label{table:pos_encoding}
\end{table}

\begin{table}[!b]
	\centering
	\caption{Ablation study of the random and the sequential sampling in RV-based ball query.}
	\vspace{-1.0em}
	\resizebox{0.5\columnwidth}!{\begin{tabular}{c|cc}
		\toprule
		Methods & Random & Sequential \\
		\midrule
		APH & \textbf{65.01} & 64.11 \\
		\bottomrule
	\end{tabular}}
	\label{table:random_seq_ball_query}
\end{table}

\begin{figure}[!t]
	\centering
    \vspace{0.2em}
    \includegraphics[width=0.8\columnwidth]{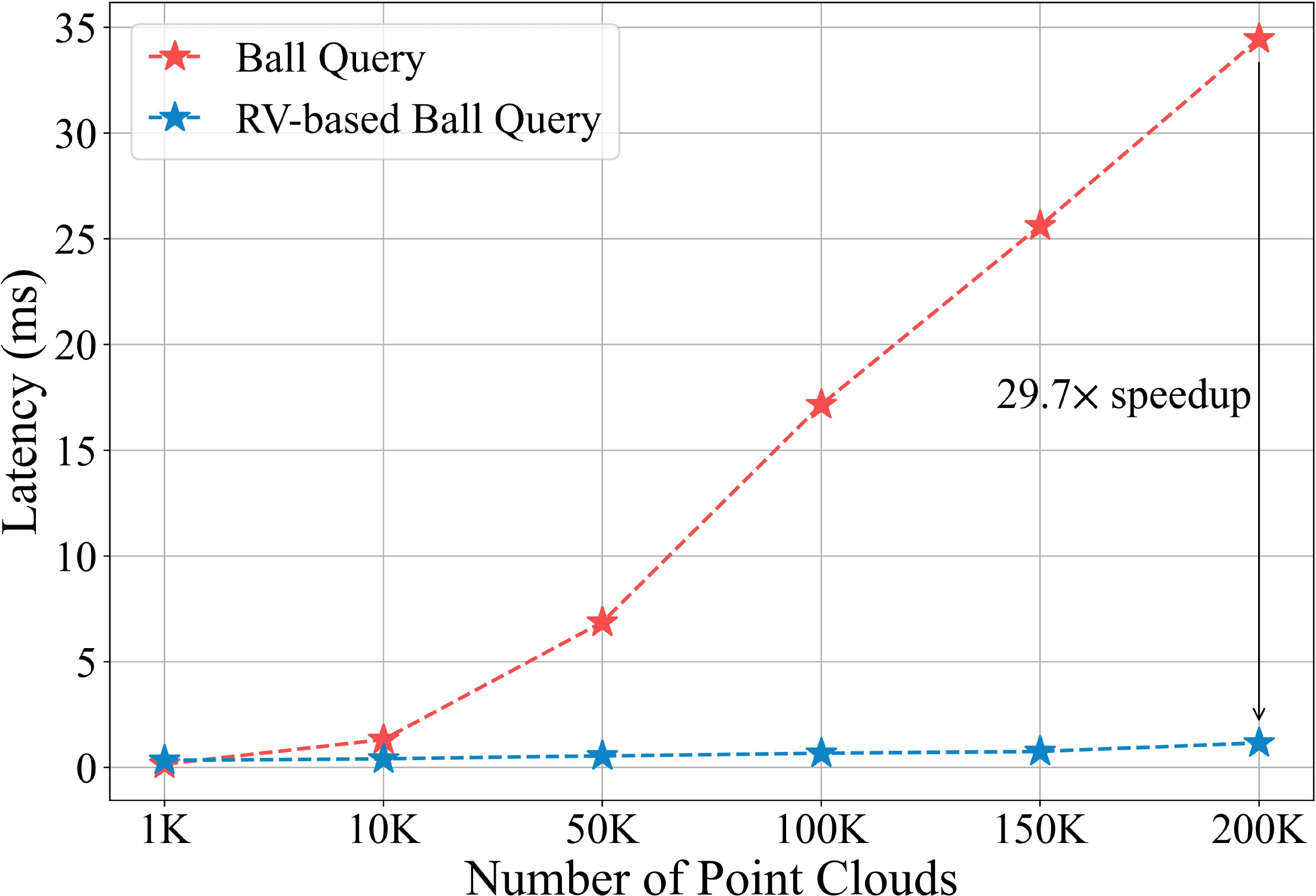}
    \vspace{-1.0em}
    \caption{Comparisons between range-view-based ball query and original ball query.}
	\label{fig:bq_statis}
	\vspace{-1.8em}
\end{figure}

\textbf{Virtual Range Image.}
Table~\ref{table:random_seq_ball_query} illustrates the comparison between random and sequential sampling used in the range-view-based ball query. 
Sequential sampling means that we traverse the kernel in order from top-left to bottom-right and sample points within the radius until the number of points is satisfied.
We observe an APH drop of 0.9 when using sequential sampling since it may result in an uneven distribution of sampled points, i.e., points in the top-left part are more likely to be sampled.
Figure~\ref{fig:bq_statis} compares the latency of our range-view-based ball with that of the original ball query, where the kernel size is set to $16$ to find $32$ neighbors within a radius of $0.8$.
It shows that the range view-based ball query performs well as the point cloud increases and achieves a speedup of 29.7$\times$ over the original ball query when it is applied on 200K point clouds.

\begin{figure}[!h]
	\centering
        \vspace{-0.6em}
	\includegraphics[width=1.0\columnwidth]{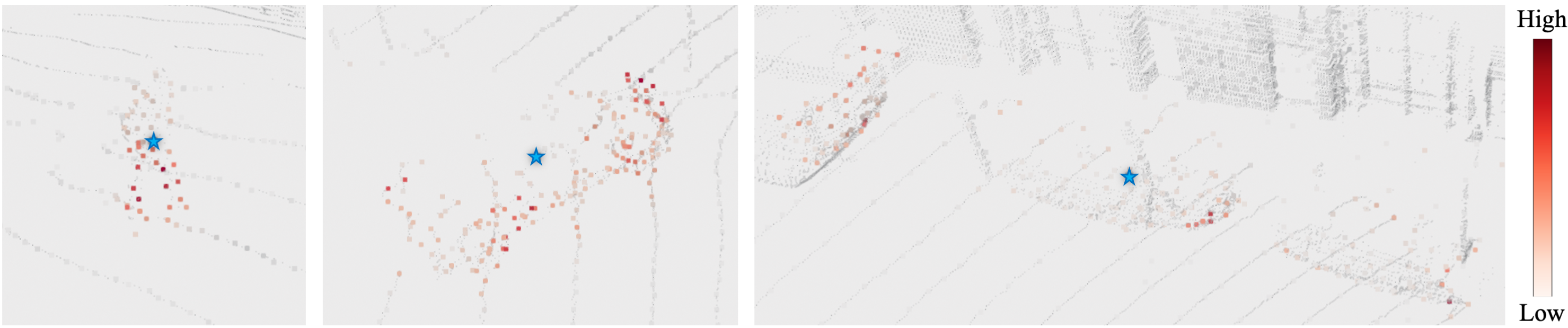}
	\vspace{-2.2em}
	\caption{Visualization of the attention map.}
	\label{fig:attn_vis}
        \vspace{-1.6em}
\end{figure}

\textbf{Visualization.}
Figure~\ref{fig:attn_vis} shows the attention weights of the sampled tokens for each reference point.
The model can capture object-centric features and long-range features.

\section{Conclusion}

PVT-SSD leverages the benefits from the voxel and point representations.
We propose the Query Initialization module to generate reference points and content queries to associate these two different representations.
Then, the Point-Voxel Transformer module is introduced to capture long-range contextual features from voxels and fine-grained geometric features from points.
To accelerate the neighbor querying process, we design a Virtual Range Image module.
The constructed range image is a generalized version of the native range image captured by LiDAR sensors and thus can be used for more scenarios.
Experiments on several autonomous driving benchmarks demonstrate the efficacy.


\vspace{-6pt}
\paragraph{Acknowledgement}
This work was supported in part by The National Nature Science Foundation of China (Grant Nos:  U1909203, 62273303), in part by Innovation Capability Support Program of Shaanxi (Program No. 2021TD-05), in part by the Key R\&D Program of Zhejiang Province, China (2023C01135), in part by S\&T Plan of Zhejiang Province (No. 202218), in part by the National Key R\&D Program of China (No. 2022ZD0160100), and in part by Shanghai Committee of Science and Technology (Grant No. 21DZ1100100).

{\small
\bibliographystyle{ieee_fullname}
\bibliography{egbib}

\begin{thebibliography}{10}\itemsep=-1pt

\bibitem{ba2016ln}
Lei~Jimmy Ba, Jamie~Ryan Kiros, and Geoffrey~E. Hinton.
\newblock Layer normalization.
\newblock {\em CoRR}, abs/1607.06450, 2016.

\bibitem{alex2020rcd}
Alex Bewley, Pei Sun, Thomas Mensink, Drago Anguelov, and Cristian
  Sminchisescu.
\newblock Range conditioned dilated convolutions for scale invariant 3d object
  detection.
\newblock In {\em Conference on Robot Learning}, 2020.

\bibitem{caesar2020nuscenes}
Holger Caesar, Varun Bankiti, Alex~H. Lang, Sourabh Vora, Venice~Erin Liong,
  Qiang Xu, Anush Krishnan, Yu Pan, Giancarlo Baldan, and Oscar Beijbom.
\newblock nuscenes: A multimodal dataset for autonomous driving.
\newblock In {\em Proceedings of the IEEE Conference on Computer Vision and
  Pattern Recognition}, 2020.

\bibitem{carion2020detr}
Nicolas Carion, Francisco Massa, Gabriel Synnaeve, Nicolas Usunier, Alexander
  Kirillov, and Sergey Zagoruyko.
\newblock End-to-end object detection with transformers.
\newblock In {\em Proceedings of the European Conference on Computer Vision},
  2020.

\bibitem{chai2021ppc}
Yuning Chai, Pei Sun, Jiquan Ngiam, Weiyue Wang, Benjamin Caine, Vijay
  Vasudevan, Xiao Zhang, and Dragomir Anguelov.
\newblock To the point: Efficient 3d object detection in the range image with
  graph convolution kernels.
\newblock In {\em Proceedings of the IEEE Conference on Computer Vision and
  Pattern Recognition}, 2021.

\bibitem{chen2022sasa}
Chen Chen, Zhe Chen, Jing Zhang, and Dacheng Tao.
\newblock {SASA:} semantics-augmented set abstraction for point-based 3d object
  detection.
\newblock In {\em Proceedings of the AAAI Conference on Artificial
  Intelligence}, 2022.

\bibitem{deng2021voxelrcnn}
Jiajun Deng, Shaoshuai Shi, Peiwei Li, Wengang Zhou, Yanyong Zhang, and
  Houqiang Li.
\newblock Voxel r-cnn: Towards high performance voxel-based 3d object
  detection.
\newblock In {\em Proceedings of the AAAI Conference on Artificial
  Intelligence}, 2021.

\bibitem{deng2022vista}
Shengheng Deng, Zhihao Liang, Lin Sun, and Kui Jia.
\newblock Vista: Boosting 3d object detection via dual cross-view spatial
  attention.
\newblock In {\em Proceedings of the IEEE Conference on Computer Vision and
  Pattern Recognition}, 2022.

\bibitem{dong2022mssvt}
Shaocong Dong, Lihe Ding, Haiyang Wang, Tingfa Xu, Xinli Xu, Jie Wang, Ziyang
  Bian, Ying Wang, and Jianan Li.
\newblock Mssvt: Mixed-scale sparse voxel transformer for 3d object detection
  on point clouds.
\newblock In {\em Advances in Neural Information Processing Systems}, 2022.

\bibitem{fan2022sst}
Lue Fan, Ziqi Pang, Tianyuan Zhang, Yu{-}Xiong Wang, Hang Zhao, Feng Wang,
  Naiyan Wang, and Zhaoxiang Zhang.
\newblock Embracing single stride 3d object detector with sparse transformer.
\newblock In {\em Proceedings of the IEEE Conference on Computer Vision and
  Pattern Recognition}, 2022.

\bibitem{fan2022fsd}
Lue Fan, Feng Wang, Naiyan Wang, and Zhaoxiang Zhang.
\newblock Fully sparse 3d object detection.
\newblock In {\em Advances in Neural Information Processing Systems}, 2022.

\bibitem{fan2021rangedet}
Lue Fan, Xuan Xiong, Feng Wang, Naiyan Wang, and Zhaoxiang Zhang.
\newblock Rangedet: In defense of range view for lidar-based 3d object
  detection.
\newblock In {\em Proceedings of the IEEE International Conference on Computer
  Vision}, 2021.

\bibitem{geiger2012kitti}
Andreas Geiger, Philip Lenz, and Raquel Urtasun.
\newblock Are we ready for autonomous driving? the {KITTI} vision benchmark
  suite.
\newblock In {\em Proceedings of the IEEE Conference on Computer Vision and
  Pattern Recognition}, 2012.

\bibitem{guan2022m3detr}
Tianrui Guan, Jun Wang, Shiyi Lan, Rohan Chandra, Zuxuan Wu, Larry Davis, and
  Dinesh Manocha.
\newblock {M3DETR:} multi-representation, multi-scale, mutual-relation 3d
  object detection with transformers.
\newblock In {\em Winter Conference on Applications of Computer Vision}, 2022.

\bibitem{he2022voxset}
Chenhang He, Ruihuang Li, Shuai Li, and Lei Zhang.
\newblock Voxel set transformer: {A} set-to-set approach to 3d object detection
  from point clouds.
\newblock In {\em Proceedings of the IEEE Conference on Computer Vision and
  Pattern Recognition}, 2022.

\bibitem{he2020sassd}
Chenhang He, Hui Zeng, Jianqiang Huang, Xian-Sheng Hua, and Lei Zhang.
\newblock Structure aware single-stage 3d object detection from point cloud.
\newblock In {\em Proceedings of the IEEE Conference on Computer Vision and
  Pattern Recognition}, 2020.

\bibitem{he2022svganet}
Qingdong He, Zhengning Wang, Hao Zeng, Yi Zeng, and Yijun Liu.
\newblock Svga-net: Sparse voxel-graph attention network for 3d object
  detection from point clouds.
\newblock In {\em Proceedings of the AAAI Conference on Artificial
  Intelligence}, 2022.

\bibitem{hu2022pdv}
Jordan S.~K. Hu, Tianshu Kuai, and Steven~L. Waslander.
\newblock Point density-aware voxels for lidar 3d object detection.
\newblock In {\em Proceedings of the IEEE Conference on Computer Vision and
  Pattern Recognition}, 2022.

\bibitem{hu2020randlanet}
Qingyong Hu, Bo Yang, Linhai Xie, Stefano Rosa, Yulan Guo, Zhihua Wang, Niki
  Trigoni, and Andrew Markham.
\newblock Randla-net: Efficient semantic segmentation of large-scale point
  clouds.
\newblock In {\em Proceedings of the IEEE Conference on Computer Vision and
  Pattern Recognition}, 2020.

\bibitem{lai2022stratified}
Xin Lai, Jianhui Liu, Li Jiang, Liwei Wang, Hengshuang Zhao, Shu Liu, Xiaojuan
  Qi, and Jiaya Jia.
\newblock Stratified transformer for 3d point cloud segmentation.
\newblock In {\em Proceedings of the IEEE Conference on Computer Vision and
  Pattern Recognition}, 2022.

\bibitem{lang2019pointpillar}
Alex~H. Lang, Sourabh Vora, Holger Caesar, Lubing Zhou, Jiong Yang, and Oscar
  Beijbom.
\newblock Pointpillars: Fast encoders for object detection from point clouds.
\newblock In {\em Proceedings of the IEEE Conference on Computer Vision and
  Pattern Recognition}, 2019.

\bibitem{li2022deepfusion}
Yingwei Li, Adams~Wei Yu, Tianjian Meng, Benjamin Caine, Jiquan Ngiam, Daiyi
  Peng, Junyang Shen, Bo Wu, Yifeng Lu, Denny Zhou, Quoc~V. Le, Alan~L. Yuille,
  and Mingxing Tan.
\newblock Deepfusion: Lidar-camera deep fusion for multi-modal 3d object
  detection.
\newblock In {\em Proceedings of the IEEE Conference on Computer Vision and
  Pattern Recognition}, 2022.

\bibitem{li2021lidarrcnn}
Zhichao Li, Feng Wang, and Naiyan Wang.
\newblock Lidar r-cnn: An efficient and universal 3d object detector.
\newblock In {\em Proceedings of the IEEE Conference on Computer Vision and
  Pattern Recognition}, 2021.

\bibitem{liang2021rangeioudet}
Zhidong Liang, Zehan Zhang, Ming Zhang, Xian Zhao, and Shiliang Pu.
\newblock Rangeioudet: Range image based real-time 3d object detector optimized
  by intersection over union.
\newblock In {\em Proceedings of the IEEE Conference on Computer Vision and
  Pattern Recognition}, 2021.

\bibitem{liu20223dqueryis}
Jiaheng Liu, Tong He, Honghui Yang, Rui Su, Jiayi Tian, Junran Wu, Hongcheng
  Guo, Ke Xu, and Wanli Ouyang.
\newblock 3d-queryis: {A} query-based framework for 3d instance segmentation.
\newblock {\em CoRR}, abs/2211.09375, 2022.

\bibitem{liu2021swin}
Ze Liu, Yutong Lin, Yue Cao, Han Hu, Yixuan Wei, Zheng Zhang, Stephen Lin, and
  Baining Guo.
\newblock Swin transformer: Hierarchical vision transformer using shifted
  windows.
\newblock In {\em Proceedings of the IEEE International Conference on Computer
  Vision}, 2021.

\bibitem{liu2019pvcnn}
Zhijian Liu, Haotian Tang, Yujun Lin, and Song Han.
\newblock Point-voxel {CNN} for efficient 3d deep learning.
\newblock In Hanna~M. Wallach, Hugo Larochelle, Alina Beygelzimer, Florence
  d'Alch{\'{e}}{-}Buc, Emily~B. Fox, and Roman Garnett, editors, {\em Advances
  in Neural Information Processing Systems}, 2019.

\bibitem{liu2021sparsepoint}
Zili Liu, Guodong Xu, Honghui Yang, Minghao Chen, Kuoliang Wu, Zheng Yang,
  Haifeng Liu, and Deng Cai.
\newblock Suppress-and-refine framework for end-to-end 3d object detection.
\newblock {\em CoRR}, abs/2103.10042, 2021.

\bibitem{liu2021groupfree}
Ze Liu, Zheng Zhang, Yue Cao, Han Hu, and Xin Tong.
\newblock Group-free 3d object detection via transformers.
\newblock In {\em Proceedings of the IEEE International Conference on Computer
  Vision}, 2021.

\bibitem{liu2019tanet}
Zhe Liu, Xin Zhao, Tengteng Huang, Ruolan Hu, Yu Zhou, and Xiang Bai.
\newblock Tanet: Robust 3d object detection from point clouds with triple
  attention.
\newblock In {\em Proceedings of the AAAI Conference on Artificial
  Intelligence}, 2020.

\bibitem{losh2019adamw}
Ilya Loshchilov and Frank Hutter.
\newblock Decoupled weight decay regularization.
\newblock In {\em International Conference on Learning Representations}, 2019.

\bibitem{mao2021pyramid}
Jiageng Mao, Minzhe Niu, Haoyue Bai, Xiaodan Liang, Hang Xu, and Chunjing Xu.
\newblock Pyramid r-cnn: Towards better performance and adaptability for 3d
  object detection.
\newblock In {\em Proceedings of the IEEE International Conference on Computer
  Vision}, 2021.

\bibitem{mao2021votr}
Jiageng Mao, Yujing Xue, Minzhe Niu, Haoyue Bai, Jiashi Feng, Xiaodan Liang,
  Hang Xu, and Chunjing Xu.
\newblock Voxel transformer for 3d object detection.
\newblock In {\em Proceedings of the IEEE International Conference on Computer
  Vision}, 2021.

\bibitem{meyer2019lasernet}
Gregory~P. Meyer, Ankit Laddha, Eric Kee, Carlos Vallespi-Gonzalez, and Carl~K.
  Wellington.
\newblock Lasernet: An efficient probabilistic 3d object detector for
  autonomous driving.
\newblock In {\em Proceedings of the IEEE Conference on Computer Vision and
  Pattern Recognition}, 2019.

\bibitem{miao2021pvgnet}
Zhenwei Miao, Jikai Chen, Hongyu Pan, Ruiwen Zhang, Kaixuan Liu, Peihan Hao,
  Jun Zhu, Yang Wang, and Xin Zhan.
\newblock Pvgnet: {A} bottom-up one-stage 3d object detector with integrated
  multi-level features.
\newblock In {\em Proceedings of the IEEE Conference on Computer Vision and
  Pattern Recognition}, 2021.

\bibitem{misra20213detr}
Ishan Misra, Rohit Girdhar, and Armand Joulin.
\newblock An end-to-end transformer model for 3d object detection.
\newblock In {\em Proceedings of the IEEE International Conference on Computer
  Vision}, 2021.

\bibitem{pan2021pointformer}
Xuran Pan, Zhuofan Xia, Shiji Song, Li~Erran Li, and Gao Huang.
\newblock 3d object detection with pointformer.
\newblock In {\em Proceedings of the IEEE Conference on Computer Vision and
  Pattern Recognition}, 2021.

\bibitem{park2022fastpointtransformer}
Chunghyun Park, Yoonwoo Jeong, Minsu Cho, and Jaesik Park.
\newblock Fast point transformer.
\newblock In {\em Proceedings of the IEEE Conference on Computer Vision and
  Pattern Recognition}, 2022.

\bibitem{qi2019votenet}
Charles~R. Qi, Or Litany, Kaiming He, and Leonidas~J. Guibas.
\newblock Deep hough voting for 3d object detection in point clouds.
\newblock In {\em Proceedings of the IEEE International Conference on Computer
  Vision}, 2019.

\bibitem{qi2017pointnet}
Charles~R. Qi, Hao Su, Kaichun Mo, and Leonidas~J. Guibas.
\newblock Pointnet: Deep learning on point sets for 3d classification and
  segmentation.
\newblock In {\em Proceedings of the IEEE Conference on Computer Vision and
  Pattern Recognition}, 2017.

\bibitem{qi2017pointnet++}
Charles~Ruizhongtai Qi, Li Yi, Hao Su, and Leonidas~J. Guibas.
\newblock Pointnet++: Deep hierarchical feature learning on point sets in a
  metric space.
\newblock In {\em Advances in Neural Information Processing Systems}, 2017.

\bibitem{sheng2021ct3d}
Hualian Sheng, Sijia Cai, Yuan Liu, Bing Deng, Jianqiang Huang, Xian-Sheng Hua,
  and Min-Jian Zhao.
\newblock Improving 3d object detection with channel-wise transformer.
\newblock In {\em Proceedings of the IEEE International Conference on Computer
  Vision}, 2021.

\bibitem{shi2020pvrcnn}
Shaoshuai Shi, Chaoxu Guo, Li Jiang, Zhe Wang, Jianping Shi, Xiaogang Wang, and
  Hongsheng Li.
\newblock Pv-rcnn: Point-voxel feature set abstraction for 3d object detection.
\newblock In {\em Proceedings of the IEEE Conference on Computer Vision and
  Pattern Recognition}, 2020.

\bibitem{shi2021pvrcnnplusplus}
Shaoshuai Shi, Li Jiang, Jiajun Deng, Zhe Wang, Chaoxu Guo, Jianping Shi,
  Xiaogang Wang, and Hongsheng Li.
\newblock {PV-RCNN++:} point-voxel feature set abstraction with local vector
  representation for 3d object detection.
\newblock {\em CoRR}, abs/2102.00463, 2021.

\bibitem{shi2019pointrcnn}
Shaoshuai Shi, Xiaogang Wang, and Hongsheng Li.
\newblock Pointrcnn: 3d object proposal generation and detection from point
  cloud.
\newblock In {\em Proceedings of the IEEE Conference on Computer Vision and
  Pattern Recognition}, 2019.

\bibitem{shi2020part}
Shaoshuai Shi, Zhe Wang, Jianping Shi, Xiaogang Wang, and Hongsheng Li.
\newblock From points to parts: 3d object detection from point cloud with
  part-aware and part-aggregation network.
\newblock {\em IEEE Transactions on Pattern Analysis and Machine Intelligence},
  2020.

\bibitem{shi2020pointgnn}
Weijing Shi and Raj Rajkumar.
\newblock Point-gnn: Graph neural network for 3d object detection in a point
  cloud.
\newblock In {\em Proceedings of the IEEE Conference on Computer Vision and
  Pattern Recognition}, 2020.

\bibitem{sun2020wod}
Pei Sun, Henrik Kretzschmar, Xerxes Dotiwalla, Aurelien Chouard, Vijaysai
  Patnaik, Paul Tsui, James Guo, Yin Zhou, Yuning Chai, Benjamin Caine, Vijay
  Vasudevan, Wei Han, Jiquan Ngiam, Hang Zhao, Aleksei Timofeev, Scott
  Ettinger, Maxim Krivokon, Amy Gao, Aditya Joshi, Yu Zhang, Jonathon Shlens,
  Zhifeng Chen, and Dragomir Anguelov.
\newblock Scalability in perception for autonomous driving: Waymo open dataset.
\newblock In {\em Proceedings of the IEEE Conference on Computer Vision and
  Pattern Recognition}, 2020.

\bibitem{sun2022swformer}
Pei Sun, Mingxing Tan, Weiyue Wang, Chenxi Liu, Fei Xia, Zhaoqi Leng, and
  Dragomir Anguelov.
\newblock Swformer: Sparse window transformer for 3d object detection in point
  clouds.
\newblock In {\em Proceedings of the European Conference on Computer Vision},
  2022.

\bibitem{sun2021rsn}
Pei Sun, Weiyue Wang, Yuning Chai, Gamaleldin Elsayed, Alex Bewley, Xiao Zhang,
  Cristian Sminchisescu, and Dragomir Anguelov.
\newblock {RSN:} range sparse net for efficient, accurate lidar 3d object
  detection.
\newblock In {\em Proceedings of the IEEE Conference on Computer Vision and
  Pattern Recognition}, 2021.

\bibitem{ashish2017transformer}
Ashish Vaswani, Noam Shazeer, Niki Parmar, Jakob Uszkoreit, Llion Jones,
  Aidan~N. Gomez, Lukasz Kaiser, and Illia Polosukhin.
\newblock Attention is all you need.
\newblock In {\em Advances in Neural Information Processing Systems}, 2017.

\bibitem{wang2022rbgnet}
Haiyang Wang, Shaoshuai Shi, Ze Yang, Rongyao Fang, Qi Qian, Hongsheng Li,
  Bernt Schiele, and Liwei Wang.
\newblock Rbgnet: Ray-based grouping for 3d object detection.
\newblock In {\em Proceedings of the IEEE Conference on Computer Vision and
  Pattern Recognition}, 2022.

\bibitem{wang2020piilar-od}
Yue Wang, Alireza Fathi, Abhijit Kundu, David~A. Ross, Caroline Pantofaru,
  Thomas~A. Funkhouser, and Justin Solomon.
\newblock Pillar-based object detection for autonomous driving.
\newblock In {\em Proceedings of the European Conference on Computer Vision},
  2020.

\bibitem{wang2021objectdgcnn}
Yue Wang and Justin Solomon.
\newblock Object {DGCNN:} 3d object detection using dynamic graphs.
\newblock In {\em Advances in Neural Information Processing Systems}, 2021.

\bibitem{wu2021crpe}
Kan Wu, Houwen Peng, Minghao Chen, Jianlong Fu, and Hongyang Chao.
\newblock Rethinking and improving relative position encoding for vision
  transformer.
\newblock In {\em Proceedings of the IEEE International Conference on Computer
  Vision}, 2021.

\bibitem{wu2022sfd}
Xiaopei Wu, Liang Peng, Honghui Yang, Liang Xie, Chenxi Huang, Chengqi Deng,
  Haifeng Liu, and Deng Cai.
\newblock Sparse fuse dense: Towards high quality 3d detection with depth
  completion.
\newblock In {\em Proceedings of the IEEE Conference on Computer Vision and
  Pattern Recognition}, 2022.

\bibitem{xu2022btcdet}
Qiangeng Xu, Yiqi Zhong, and Ulrich Neumann.
\newblock Behind the curtain: Learning occluded shapes for 3d object detection.
\newblock In {\em Proceedings of the AAAI Conference on Artificial
  Intelligence}, 2022.

\bibitem{xv2021spg}
Qiangeng Xu, Yin Zhou, Weiyue Wang, Charles~R. Qi, and Dragomir Anguelov.
\newblock {SPG:} unsupervised domain adaptation for 3d object detection via
  semantic point generation.
\newblock In {\em Proceedings of the IEEE International Conference on Computer
  Vision}, 2021.

\bibitem{xue2022point2seq}
Yujing Xue, Jiageng Mao, Minzhe Niu, Hang Xu, Michael~Bi Mi, Wei Zhang,
  Xiaogang Wang, and Xinchao Wang.
\newblock Point2seq: Detecting 3d objects as sequences.
\newblock In {\em Proceedings of the IEEE Conference on Computer Vision and
  Pattern Recognition}, 2022.

\bibitem{yan2018second}
Yan Yan, Yuxing Mao, and Bo Li.
\newblock Second: Sparsely embedded convolutional detection.
\newblock {\em Sensors}, 18(10), 2018.

\bibitem{Yang2018pixor}
Bin Yang, Wenjie Luo, and Raquel Urtasun.
\newblock Pixor: Real-time 3d object detection from point clouds.
\newblock In {\em Proceedings of the IEEE Conference on Computer Vision and
  Pattern Recognition}, 2018.

\bibitem{yang2023gd-mae}
Honghui Yang, Tong He, Jiaheng Liu, Hua Chen, Boxi Wu, Binbin Lin, Xiaofei He,
  and Wanli Ouyang.
\newblock {GD-MAE:} generative decoder for {MAE} pre-training on lidar point
  clouds.
\newblock In {\em Proceedings of the IEEE Conference on Computer Vision and
  Pattern Recognition}, 2023.

\bibitem{yang2022graphrcnn}
Honghui Yang, Zili Liu, Xiaopei Wu, Wenxiao Wang, Wei Qian, Xiaofei He, and
  Deng Cai.
\newblock Graph {R-CNN:} towards accurate 3d object detection with
  semantic-decorated local graph.
\newblock In {\em Proceedings of the European Conference on Computer Vision},
  2022.

\bibitem{yang2022eqpvrcnn}
Zetong Yang, Li Jiang, Yanan Sun, Bernt Schiele, and Jiaya Jia.
\newblock A unified query-based paradigm for point cloud understanding.
\newblock In {\em Proceedings of the IEEE Conference on Computer Vision and
  Pattern Recognition}, 2022.

\bibitem{yang20203dssd}
Zetong Yang, Yanan Sun, Shu Liu, and Jiaya Jia.
\newblock 3dssd: Point-based 3d single stage object detector.
\newblock In {\em Proceedings of the IEEE Conference on Computer Vision and
  Pattern Recognition}, 2020.

\bibitem{yang2019std}
Zetong Yang, Yanan Sun, Shu Liu, Xiaoyong Shen, and Jiaya Jia.
\newblock Std: Sparse-to-dense 3d object detector for point cloud.
\newblock In {\em Proceedings of the IEEE International Conference on Computer
  Vision}, 2019.

\bibitem{yang20213dman}
Zetong Yang, Yin Zhou, Zhifeng Chen, and Jiquan Ngiam.
\newblock 3d-man: 3d multi-frame attention network for object detection.
\newblock In {\em Proceedings of the IEEE Conference on Computer Vision and
  Pattern Recognition}, 2021.

\bibitem{ye2022lidarmultinet}
Dongqiangzi Ye, Weijia Chen, Zixiang Zhou, Yufei Xie, Yu Wang, Panqu Wang, and
  Hassan Foroosh.
\newblock Lidarmultinet: Unifying lidar semantic segmentation, 3d object
  detection, and panoptic segmentation in a single multi-task network.
\newblock {\em CoRR}, abs/2206.11428, 2022.

\bibitem{ye2020hvnet}
Maosheng Ye, Shuangjie Xu, and Tongyi Cao.
\newblock Hvnet: Hybrid voxel network for lidar based 3d object detection.
\newblock In {\em Proceedings of the IEEE Conference on Computer Vision and
  Pattern Recognition}, 2020.

\bibitem{yin2021center}
Tianwei Yin, Xingyi Zhou, and Philipp Kr{\"a}henb{\"u}hl.
\newblock Center-based 3d object detection and tracking.
\newblock In {\em Proceedings of the IEEE Conference on Computer Vision and
  Pattern Recognition}, 2021.

\bibitem{zhang2020moca}
Wenwei Zhang, Zhe Wang, and Chen~Change Loy.
\newblock Multi-modality cut and paste for 3d object detection.
\newblock {\em CoRR}, abs/2012.12741, 2020.

\bibitem{zhang2022iassd}
Yifan Zhang, Qingyong Hu, Guoquan Xu, Yanxin Ma, Jianwei Wan, and Yulan Guo.
\newblock Not all points are equal: Learning highly efficient point-based
  detectors for 3d lidar point clouds.
\newblock In {\em Proceedings of the IEEE Conference on Computer Vision and
  Pattern Recognition}, 2022.

\bibitem{zhao2021pointtransformer}
Hengshuang Zhao, Li Jiang, Jiaya Jia, Philip H.~S. Torr, and Vladlen Koltun.
\newblock Point transformer.
\newblock In {\em Proceedings of the IEEE International Conference on Computer
  Vision}, 2021.

\bibitem{zheng2020ciassd}
Wu Zheng, Weiliang Tang, Sijin Chen, Li Jiang, and Chi-Wing Fu.
\newblock Cia-ssd: Confident iou-aware single-stage object detector from point
  cloud.
\newblock In {\em Proceedings of the AAAI Conference on Artificial
  Intelligence}, 2021.

\bibitem{zheng2021sessd}
Wu Zheng, Weiliang Tang, Li Jiang, and Chi-Wing Fu.
\newblock Se-ssd: Self-ensembling single-stage object detector from point
  cloud.
\newblock In {\em Proceedings of the IEEE Conference on Computer Vision and
  Pattern Recognition}, 2021.

\bibitem{zhou2018voxelnet}
Yin Zhou and Oncel Tuzel.
\newblock Voxelnet: End-to-end learning for point cloud based 3d object
  detection.
\newblock In {\em Proceedings of the IEEE Conference on Computer Vision and
  Pattern Recognition}, 2018.

\bibitem{zhou2022centerformer}
Zixiang Zhou, Xiangchen Zhao, Yu Wang, Panqu Wang, and Hassan Foroosh.
\newblock Centerformer: Center-based transformer for 3d object detection.
\newblock In {\em Proceedings of the European Conference on Computer Vision},
  2022.

\bibitem{zhu2021deformabledetr}
Xizhou Zhu, Weijie Su, Lewei Lu, Bin Li, Xiaogang Wang, and Jifeng Dai.
\newblock Deformable {DETR:} deformable transformers for end-to-end object
  detection.
\newblock In {\em International Conference on Learning Representations}, 2021.

\end{thebibliography}
}

\end{document}